%% file: main.tex
\documentclass[a4paper,conference]{IEEEtran}

\usepackage{cite}
\usepackage{url, hyperref}

\ifCLASSOPTIONcompsoc
 \usepackage[caption=false,font=normalsize,labelfont=sf,textfont=sf]{subfig}
\else
 \usepackage[caption=false,font=footnotesize]{subfig}
\fi
 \usepackage{amsmath, amsfonts}
\usepackage{multirow, array, booktabs, tabularx, wrapfig}
\usepackage{adjustbox}
\usepackage{arydshln}
\usepackage{pifont}
\usepackage{xcolor}
\usepackage{bbold}
\usepackage{algorithm}
\usepackage[noend]{algpseudocode}
\usepackage{mathtools}
\usepackage{csquotes}

\newcommand{\defeq}{\vcentcolon=} 
\newcommand{\cmark}{\ding{51}} 
\newcommand{\xmark}{\ding{55}} 

\algnewcommand\algorithmicforeach{\textbf{for each}}
\algdef{S}[FOR]{ForEach}[1]{\algorithmicforeach\ #1\ \algorithmicdo}

\newtheorem{theorem}{Theorem}

\title{Domain Generalization via Selective Consistency Regularization for Time Series Classification}

\author{\IEEEauthorblockN{Wenyu Zhang}
\IEEEauthorblockA{Institute for Infocomm Research, A*STAR}
\and
\IEEEauthorblockN{Mohamed Ragab and Chuan-Sheng Foo}
\IEEEauthorblockA{Institute for Infocomm Research, A*STAR \\
Centre for Frontier AI Research, A*STAR}}

\begin{document}

\maketitle

\begin{abstract}
Domain generalization methods aim to learn models robust to domain shift with data from a limited number of source domains and without access to target domain samples during training. Popular domain alignment methods for domain generalization seek to extract domain-invariant features by minimizing the discrepancy between feature distributions across all domains, disregarding inter-domain relationships. In this paper, we instead propose a novel representation learning methodology that selectively enforces prediction consistency between source domains estimated to be closely-related. Specifically, we hypothesize that domains share different class-informative representations, so instead of aligning all domains which can cause negative transfer,
we only regularize the discrepancy between closely-related domains. 
We apply our method to time-series classification tasks and conduct comprehensive experiments on three public real-world datasets. Our method significantly improves over the baseline and achieves better or competitive performance in comparison with state-of-the-art methods in terms of both accuracy and model calibration.
\end{abstract}

\input{sections/introduction}
\input{sections/related_works}
\input{sections/proposed_method}

\input{sections/experiments_and_results}
\input{sections/further_analysis}
\input{sections/conclusion}

\newpage

\section*{Acknowledgments}
This research is supported by the Agency for Science, Technology and Research (A*STAR) under its AME Programmatic Funds (Grant No. A20H6b0151).

\bibliography{references}
\bibliographystyle{IEEEtran}

\clearpage

\appendix

\input{sections/appendix_implementation_details}
\input{sections/appendix_further_experiment_results}

\end{document}

%% file: sections/introduction.tex
\section{Introduction}
\label{sec: introduction}
Increasing accessibility to data has spurred the use of data-driven deep learning methods. In practical deployments, models need to be robust to data distribution shifts between training \textit{(source domains)} and test \textit{(target domain)} data, when data are collected for the same task but in different environments~\cite{gulrajani2020domainbed, Hendrycks2020robustness}. Such domain shift may occur as the collection of training data is subject to resource constraints and may not provide sufficient coverage, or is conducted in controlled settings that do not fully assimilate real environments~\cite{Li2018MLDG, Zhang2020CAZSLZR, gupta2018homes}.

In this work, we consider the \textit{domain generalization} (DG) setting where we leverage data from multiple source domains to generalize to unseen target domains. This differs from \textit{domain adaptation} (DA) which assumes access to unlabeled target samples for training. We further focus on time series classification; while there are many works studying DG in image classification~\cite{Hendrycks2020robustness, gulrajani2020domainbed,Li2019EpisodicTF, Mancini2018BestSF, somavarapu2020stylization, Li2018MLDG, fabio2019jigsaw} and natural language tasks~\cite{Ganin2016DANN, guo2018mixture, balaji2018metareg}, there is limited literature and evaluation for time series classification. We therefore aim to address this gap. Example use-cases include generalization across human subjects for activity recognition~\cite{garrett2020sensor}, and across operating conditions for sensor-based machine fault detection~\cite{zheng2020diagnosis}.
\begin{figure}[h]
    \centering
    \includegraphics[width  =0.4\textwidth]{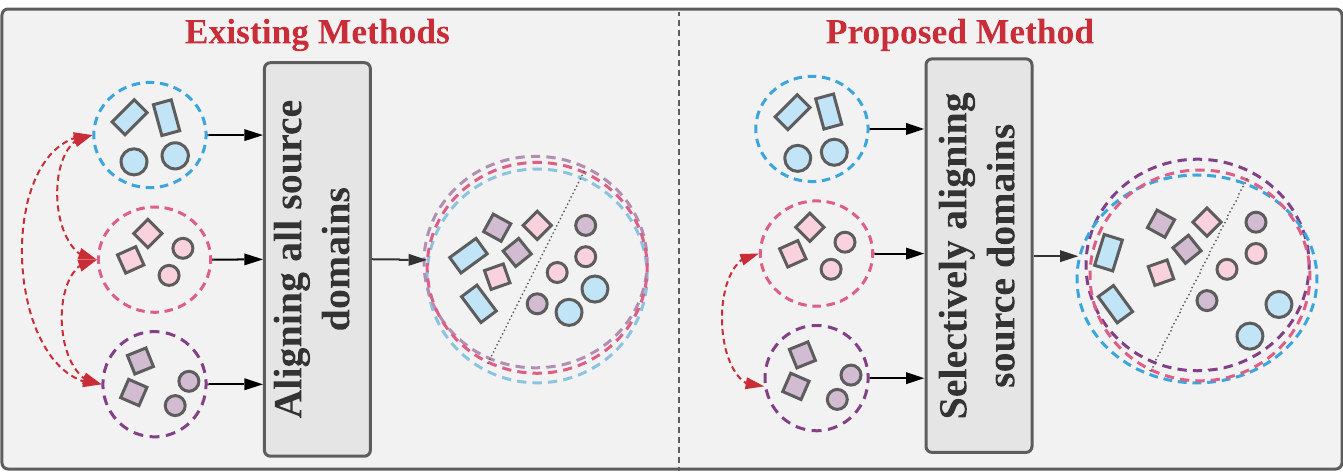}
    \vspace{-3mm}     
    \caption{\textbf{Left}: Existing domain alignment methods align all source domains equally. \textbf{Right}: Our proposed method, by considering inter-domain relationships, selectively aligns closely-related domains and allows greater diversity in output predictions. Best viewed in colour.}
    \label{fig:overview_comparison)}
    \vspace{-5mm}
\end{figure}

A popular DG approach is domain alignment, that is, enforcing feature distributions to be the same across all source domains~\cite{motiian2017unified, garrett2020sensor, Matsuura2020DomainGU, li2018mmd, Long2018ConditionalAD,gulrajani2020domainbed,Wang2021GeneralizingTU} to learn domain-invariant features that potentially generalize to unseen domains. However, some works in the related domain adaptation and transfer learning literature suggest that domains can have different informative characteristics and transferring between less related domains can inhibit the learning of these characteristics~\cite{guo2018mixture,wang2019negtransfer,Bousmalis2016DomainSN}, causing negative transfer and harming generalization.

\begin{figure*}[htb]
    \centering
    \includegraphics[width=0.7\textwidth]{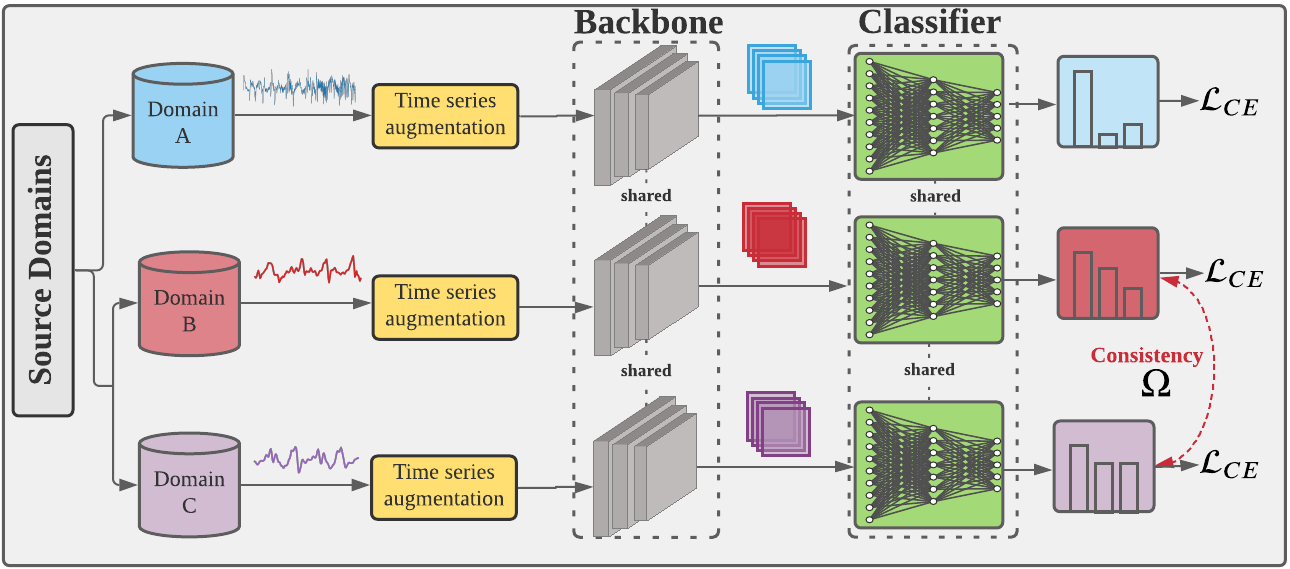}
    \vspace{-3mm}    
    \caption{Method overview. Considering multiple source domains, our proposed method enforces prediction consistency between similar domains through selective regularization of logits (pre-softmax classifier outputs). We encourage similar domains (B and C in illustration) to share predicted class relationships, while allowing diverse predicted class relationships across dissimilar domains (A vs B and C in illustration).}
    \label{fig:overview}
    \vspace{-5mm}
\end{figure*}

In this paper, we propose a novel DG approach for time series classification where we further consider cross-domain relationships during domain alignment, and regularize subsets of source domains separately based on their similarity (as illustrated in Figure~\ref{fig:overview_comparison)}).
We propose selective cross-domain consistency regularization to encourage invariant class-conditional predictions amongst similar source domains, so that model predictions are less dependent on domain-specific information, while still being allowed to be diverse to capture cross-domain differences. 
Our consistency regularization is applied on the logits (pre-softmax classifier outputs); as far we know, existing DG methods regularize on either features or soft label predictions~\cite{Ganin2016DANN, li2018mmd, Sun2016DeepCORAL, BenDavid2010domainadaptation, dou2019semantic, Matsuura2020DomainGU, gulrajani2020domainbed, Wang2018DeepVD, wang2020relations, Ahmed2021SystematicGW} or on logits in conjunction with features~\cite{kim2021selfreg}. Features are challenging to align in high-dimensional space, and soft label predictions are normalized logits and have less variability.
We leverage auxiliary domain metadata to directly infer inter-domain relationships based on application-specific knowledge when such metadata is available. Otherwise, we estimate latent relationships using a clustering-based measure of domain divergence amongst source domains. Figure~\ref{fig:overview} provides an overview of our proposed method.
We summarize our contributions as follows:
\begin{itemize}
    \item We propose a new DG methodology for time series classification that selectively enforces prediction consistency between source domains estimated to be closely-related, which helps to calibrate the model from being over and/or under-confident in its predictions. We combine this with domain-wise augmentation to generate more diverse samples in closely-related domains for regularization;
    \item The proposed method is easy to implement with data augmentation and logit regularization on top of empirical risk minimization;
    \item We perform extensive evaluation of existing methods on 3 public datasets for classification accuracy and calibration, and our proposed method demonstrates better or competitive performance compared to state-of-the-art methods.  
\end{itemize}

%% file: sections/related_works.tex
\section{Related Works}
\label{sec:related_works}

We discuss related works to learn more robust representations with a single model. Ensembles~\cite{Mancini2018BestSF, vinyals2016matching, innocente2019domain, guo2018mixture} may not scale to multiple source domains and not discussed here. 

\textbf{Data augmentation and generation:} Enriching training sample diversity naturally helps generalization. \cite{volpi2018adversarial, Shankar2018crossgrad, stutz2019disentangling} generates samples by adversarial perturbation, and \cite{Yan2020Mixup, wang2020mixup} interpolate between samples of different domains. Due to advancements in style transfer~\cite{jing2020styletransfer} and generative adversarial networks~\cite{goodfellow2014gan}, many image augmentation techniques are developed~\cite{Hendrycks2020robustness, zhou2020novel, Nam2019SagNet, gong2019dlow}. However, advanced techniques for image augmentation may not readily apply to time series.

\textbf{Learning domain-invariant features:} An approach to learn invariant features is to train the feature extractor such that the same classifier is optimal for all domains~\cite{Arjovsky2019IRM,Rosenfeld2020risksIRM}. 
Many works learn invariant features by aligning representations from all source domains. 
Some works directly minimize the distance between features or soft labels by a distance measure or adversarial networks~\cite{motiian2017unified, garrett2020sensor, Matsuura2020DomainGU, li2018mmd, Long2018ConditionalAD, albuquerque2020distmatch,mahajan2020causal, Arpit2019PredictingWH}, and some others use meta-learning to simulate domain shift~\cite{balaji2018metareg, dou2019semantic, Li2018MLDG}.
However, distribution alignment between less related domains can cause negative transfer~\cite{guo2018mixture,wang2019negtransfer,Bousmalis2016DomainSN}.

\textbf{Robustness:} Models can learn to be more robust to domain shifts through perturbations during training. \cite{Li2019EpisodicTF} episodically switches the feature extractor or classifier to domain-specific counterparts, and \cite{Huang2020RSC} zeros out features associated with the highest gradient in the classification layer to learn more diverse features.
Another approach to robustness minimizes worst-case risk over all domains, resulting in a training objective that is a weighted average of source domain losses~\cite{Sagawa2019GroupDRO, Krueger2020VREx}. 

%% file: sections/proposed_method.tex
\section{Proposed Method}
\label{sec: proposed_method}

\subsection{Preliminaries: Notation}
\label{sec: notations}

We denote total $N$ observed samples from $M$ source domains as $\left\lbrace \left(\boldsymbol{x}_n, \boldsymbol{y}_n, d_n \right) \right\rbrace_{n=1}^{N}$, where for the $n$-th sample, $\boldsymbol{x}_n$ and $\boldsymbol{y}_n$ are predictor and response, and $d_n\in \{1,\dots,M\}$ is domain label and is unavailable at test-time. $\boldsymbol{y}_n$ is a one-hot vector of the true class label in $L$ classes. Samples in each domain $d$ are independently and identically distributed according to a domain-dependent data distribution as $\left(\boldsymbol{x}, \boldsymbol{y}\right) \sim P_d(\mathcal{X},\mathcal{Y})$. 

A neural network model is composed of feature extractor $f(\cdot)$ parameterized by $\Theta$ that yields learned features $\boldsymbol{z}=f(\boldsymbol{x};\Theta)$, and classifier $h(\cdot)$ parameterized by $\Psi$ that yields logits output $\boldsymbol{g}=h(\boldsymbol{z};\Psi)$. Soft labels (a vector of estimated class probabilities) are obtained by $\boldsymbol{s}=softmax(\boldsymbol{g})$ where $\boldsymbol{s}[i] = \frac{\exp(\boldsymbol{z}[i])}{\sum_{\ell=1}^L \exp{(\boldsymbol{z}[\ell])}}$. The final predicted class is the one with the highest probability in $\boldsymbol{s}$ i.e. $ \arg\max_i \boldsymbol{s}[i]$. 

\subsection{Selective Cross-Domain Consistency Regularization}
\label{sec: selective_consistency_regularization}

We hypothesize that domains can have different informative characteristics and consequently class relationships which do not exactly match. That is, there exists domain $d^{(i)}$ and $d^{(j)}$ with $P_i(\mathcal{X}) \neq P_j(\mathcal{X})$ where there exists sample $\boldsymbol{x}^{(d^{(i)})}$ such that $P_i(\mathcal{Y}|\boldsymbol{x}^{(d^{(i)})}) \neq P_j(\mathcal{Y}|\boldsymbol{x}^{(d^{(j)})})$ for all $\boldsymbol{x}^{(d^{(j)})}$ under ground truth distributions; note $P_i(\mathcal{Y}|\boldsymbol{x}) = P_j(\mathcal{Y}|\boldsymbol{x})$ for the same $\boldsymbol{x}$. Aligning these two domains can result in inaccurate class relationships.
Specifically, objectives such as DANN~\cite{Ganin2016DANN} align $d^{(j)}$ to $d^{(i)}$ assuming that for every $\boldsymbol{x}^{(d^{(i)})}$, there exists a matching $\boldsymbol{x}^{(d^{(j)})}$ where $P_i(\mathcal{Y}|\boldsymbol{x}^{(d^{(i)})}) = P(\mathcal{Y}|f(\boldsymbol{x}^{(d^{(i)})};\Theta)) = P(\mathcal{Y}|f(\boldsymbol{x}^{(d^{(j)})};\Theta)) = P_j(\mathcal{Y}|\boldsymbol{x}^{(d^{(j)})})$. When this assumption is invalid, aligning implies that a $\boldsymbol{x}^{(d^{(j)})}$ 
where $P_i(\mathcal{Y}|\boldsymbol{x}^{(d^{(i)})}) \neq P_j(\mathcal{Y}|\boldsymbol{x}^{(d^{(j)})})$ will be aligned to  $\boldsymbol{x}^{(d^{(i)})}$ in the feature space, resulting in inaccurate class relationships since $P_i(\mathcal{Y}|\boldsymbol{x}^{(d^{(i)})}) = P(\mathcal{Y}|f(\boldsymbol{x}^{(d^{(i)})};\Theta)) = P(\mathcal{Y}|f(\boldsymbol{x}^{(d^{(j)})};\Theta)) \neq P_j(\mathcal{Y}|\boldsymbol{x}^{(d^{(j)})})$ \cite{wang2019negtransfer}.
Instead, we align only similar domains to balance between domain-invariance and diversity. We align logits instead of features to directly calibrate task-specific predictions through both feature extractor and classifier.

More formally, we aim to learn model parameters such that the conditional distribution $P(\mathcal{G} | \mathcal{Y}) = P(h(f(\mathcal{X}; \Theta); \Psi) | \mathcal{Y})$ is invariant for closely-related domains according to latent inter-domain relationships.
By encouraging invariant class-conditional predictions, model predictions can be less reliant on domain-specific information and more robust to domain shifts at test-time.
Specifically, for closely-related domains $d^{(i)}$ and $d^{(j)}$ with $\boldsymbol{g}^{(d^{(i)},\ell)} \sim P_i(\mathcal{G} | \mathcal{Y}=\ell)$ and $\boldsymbol{g}^{(d^{(j)},\ell)} \sim P_j(\mathcal{G} | \mathcal{Y}=\ell)$ for class $\ell$, we apply squared maximum mean discrepancy (MMD)~\cite{gretton2012kernel} for distribution matching. Squared MMD between two distributions $\mathbb{Q}_i$ and $\mathbb{Q}_j$ is defined as 
\begin{equation}
    MMD^2(\mathbb{Q}_i,\mathbb{Q}_j) = \| E_{\mathbb{Q}_i}(\phi(\mathcal{G})) - E_{\mathbb{Q}_j}(\phi(\mathcal{G})) \|^2_{\mathcal{B}}
\end{equation}
for reproducing kernel Hilbert space $\mathcal{B}$ and mapping function $\phi$. We take $\mathcal{B}$ as Euclidean space and $\phi$ as identity function, for which MMD is equivalent to the $L_2$ distance between distribution means. This formulation is computationally simple and empirically outperforms a more complex alternative (see Appendix B). 

The objective function for model learning is
\begin{align}
    L(\Theta,\Psi) &= L_{CE}(\Theta,\Psi) + \lambda\Omega(\Theta,\Psi) \label{eqn: overall_objective}
\end{align}
where $L_{CE}(\Theta,\Psi) = -\sum_{n=1}^N \boldsymbol{y}_n \cdot \log(\boldsymbol{s}_n)$ is the supervised cross-entropy loss, and $\Omega(\Theta,\Psi)$ is the proposed regularizer:
\begin{equation}
    \Omega(\Theta,\Psi) = \sum_{d^{(i)}=1}^M \sum_{d^{(j)}=1}^M w(d^{(i)}, d^{(j)}) \sum_{\ell=1}^L  \|\bar{\boldsymbol{g}}^{(d^{(i)},\ell)} - \bar{\boldsymbol{g}}^{(d^{(j)},\ell)}\|_2^2
    \label{eqn: regularization}
\end{equation}
where $\bar{\boldsymbol{g}}^{(d^{(i)},\ell)}$ is the mean logit vector for domain $d^{(i)}$ class $\ell$, referred to as the class-conditional domain centroid. Weights $w(d^{(i)}, d^{(j)})\geq 0$ depend on pairwise domain similarity between $d^{(i)}$ and $d^{(j)}$ to impose greater regularization on more similar domains. We describe estimation procedures for $w(d^{(i)}, d^{(j)})$ in Section~\ref{sec: estimation_latent_domain_relationships}.
The learned model is encouraged to preserve prediction consistency between similar domains. This helps to prevent the model from being over-confident in its predictions due to overfitting on spurious features of individual domains, thus improving calibration, in addition to increasing robustness to domain shifts. We summarize the proposed method in Algorithm~\ref{alg}.

\begin{algorithm}[h]
\caption{Proposed method \label{alg}}

\textbf{Input:} Training data $\left(\boldsymbol{x}, \boldsymbol{y}, d \right)$; regularization hyperparameter $\lambda$, model parameterized by $(\Theta, \Psi)$; learning rate $\eta$; \# iterations $I$; cluster update interval $U$ \\
\textbf{Additional input for metadata sim.:} Metadata $T$ \\
\textbf{Additional input for learned sim.:} RBF kernel parameter $\xi$
\begin{algorithmic}[1]

\For{iteration $i=0:I-1$}
\State Augment time series input $x \leftarrow \text{Augment}(x)$

\If{$(i \text{ mod } U)=0$}
\State Update regularization weights $w$ by Eqn~\ref{eqn: weight_fixed} (metadata sim.) or \ref{eqn: weight_learned} (learned sim.)
\EndIf

\State Get regularization $\Omega(\Theta, \Psi)$ by Eqn~\ref{eqn: regularization}
\State Get overall objective $L(\Theta, \Psi)$ by Eqn~\ref{eqn: overall_objective}
\State Update model $(\Theta,\Psi) \leftarrow (\Theta, \Psi) -\eta \nabla L(\Theta, \Psi)$
\EndFor

\end{algorithmic}
\end{algorithm}

\subsection{Defining Domain Similarity for Selective Regularization}

\label{sec: estimation_latent_domain_relationships}

Inter-domain relationships need to be determined in order to select closely-related domains for consistency regularization. We consider two scenarios: when domain metadata is available, and when it is not; domain metadata are descriptions of source domains to provide (possibly limited) context of the environments in which data is collected.

\subsubsection{Metadata Based Similarity}

With expert knowledge of the application, users can directly use available metadata to infer relationships and group the domains into clusters. Domains across clusters are assumed to not share class relationships and hence are not regularized. 
We denote the function $clust: \{1,\dots,M\} \times \{T\} \rightarrow \{1,\dots,K\}$ as the map from domain index and metadata $T$ to cluster index for $K$ clusters, and $\mathcal{S}_c=\{d | clust(d, T)=c, d\in \{1,\dots,M\}\}$ the set of domains in cluster $c$.
We set 
\begin{align}
    w_{meta}(d^{(i)}, d^{(j)}) = \sum_{c=1}^K \frac{1}{2|\mathcal{S}_c|} \mathbb{1}_{\{clust(d^{(i)},T) = clust(d^{(j)},T) = c\}} \label{eqn: weight_fixed}
\end{align}
and the consistency regularization function in Equation~\ref{eqn: regularization} can be restated as
\begin{align}
    \Omega_{meta}(\Theta,\Psi)
    = \sum_{c=1}^K \sum_{d\in \mathcal{S}_c} \sum_{\ell=1}^L \|\bar{\boldsymbol{g}}^{(d,\ell)} - \boldsymbol{\gamma}^{(c,\ell)}\|_2^2 \label{eqn: regularization_fixed}
\end{align}
where we refer to $\gamma^{(c,\ell)} = \frac{1}{|\mathcal{S}_c|} \sum_{d\in \mathcal{S}_c} \bar{\boldsymbol{g}}^{(d,\ell)}$ of cluster $c$ class $\ell$ as the class-conditional cluster centroid. 
The scaling factor in $w_{meta}$ ensures that when domain centroids are equidistant to their cluster centroids, the amount of contribution each cluster makes to the regularization is proportional to its size.

\subsubsection{Learned Similarity}

When no domain metadata is available, we propose estimating domain relationships by inter-domain divergences during training. We measure the distance between two domains for class $\ell$ as the squared $L_2$ distance between their class-conditional domain centroids, and $d^{(j)}$ is defined as the nearest neighbor domain to $d^{(i)}$ if it is nearest to $d^{(i)}$ for the most number of classes. We estimate the nearest neighbor domain as the most similar domain to $d^{(i)}$ at fixed intervals during training (every 100 iterations in our experiments), and we enforce prediction consistency between each domain and its nearest neighbor domain. That is, we set the weights $w_{learned}(d^{(i)},d^{(j)})$ as per Equation~\ref{eqn: weight_learned} if $d^{(j)}$ is the nearest neighbor and 0 otherwise, where
\begin{equation}
    w_{learned}(d^{(i)},d^{(j)}) = \frac{1}{L}\sum_{\ell=1}^L \exp\left(\frac{-\|\bar{\boldsymbol{g}}^{(d^{(i)},\ell)} - \bar{\boldsymbol{g}}^{(d^{(j)},\ell)}\|_2^2}{2\xi^2}\right)
    \label{eqn: weight_learned}
\end{equation}
by applying an RBF kernel on the inter-domain distance with hyperparameter $\xi$. We block gradients to prevent $w$ and inter-domain distance from updating in opposing directions.

\subsection{Domain-wise Time Series Augmentation}
\label{sec: time_series_augmentation}

To achieve additional robustness to data perturbations, we apply time series augmentations with 0.5 probability. For each source domain, we sample an augmentation function from a pre-defined distribution at each iteration. The domain-wise augmentation simulates potential test-time domain shifts. 
\begin{table}
\centering
\caption{Time series augmentations.
\vspace{-2mm}
\label{tab:time_series_augmentations}}
\resizebox{0.8\linewidth}{!}{
\begin{tabular}{ll}
\toprule[1pt]\midrule[0.3pt]
\multicolumn{1}{c}{\textbf{Augmentation}}   & \multicolumn{1}{c}{\textbf{General Expression}}  \\ \midrule
mean shift      & $a_{mean}(x) = x - \mu + \mu_{new}$ \\
scaling         & $a_{scale}(x) = \left(\frac{x-\mu}{\sigma}\right) * \sigma_{new} + \mu$ \\
masking         & $a_{mask}(x[i]) = 
    \begin{cases}
      x[i] & \text{w.p. } 0.9 \\
      \mu & \text{w.p. } 0.1
    \end{cases}$ \\
\midrule[0.3pt]\bottomrule[1pt]
\end{tabular}}
\vspace{-5mm}
\end{table}
We consider 3 time series augmentations, namely mean shift, scaling and masking, in Table~\ref{tab:time_series_augmentations}. The choice of augmentations depends on the dataset to avoid perturbing characteristics known to be important for classification. We provide augmentation details for each dataset in Section~\ref{sec:experiments_and_results}.

%% file: sections/experiments_and_results.tex
\section{Experiments}
\label{sec:experiments_and_results}

We compare with baseline ERM and state-of-the-art DG methods: GroupDRO~\cite{Sagawa2019GroupDRO}, VREx~\cite{Krueger2020VREx}, IRM~\cite{Arjovsky2019IRM}, Interdomain Mixup~\cite{Yan2020Mixup}, RSC~\cite{Huang2020RSC}, MTL~\cite{Blanchard2017MTL}, MLDG~\cite{Li2018MLDG} and Correlation~\cite{Arpit2019PredictingWH}. We also reformulate 4 popular domain adaption methods for DG following \cite{gulrajani2020domainbed}: MMD-DG~\cite{li2018mmd}, CORAL-DG~\cite{Sun2016DeepCORAL}, DANN-DG~\cite{Ganin2016DANN} and CDANN-DG~\cite{Li2018CDANN, Long2018ConditionalAD}.
 
We evaluate the proposed method on three real-world time series datasets, and use leave-one-domain-out evaluation following \cite{gulrajani2020domainbed}.
We provide details of the evaluation procedure, backbone networks and hyperparameters in the Appendix A. We evaluate on classification performance and expected calibration error (ECE), where the latter is a metric that measures how closely model confidence match probabilities of correct predictions and zero ECE means perfect calibration. Specifically, ECE is the expected difference between confidence and accuracy empirically approximated by
\begin{equation*}
    ECE = \sum_{j=1}^J \frac{\|B_j\|}{N} \|acc(B_j) - conf(B_j)\|
\end{equation*}
where $B_j=\{n | n \in \{1,\dots,N\}, max(\boldsymbol{s}_n) \in \left(\frac{j-1}{J}, \frac{j}{J} \right]\}$ is a bin containing indices of samples whose confidence for the predicted class falls in the corresponding interval, for $j \in \{1,\dots, J\}$ and total $J=15$ bins following \cite{guo2017calibration}. For each bin, $acc(B_j) = \frac{1}{|B_j|} \sum_{n\in B_j} \mathbb{1}(\hat{\boldsymbol{y}}_n = \boldsymbol{y}_n)$ and $conf(B_j) = \frac{1}{|B_j|} \sum_{n\in B_j} max(\boldsymbol{s}_n)$.
\input{tables/datasets}

\subsection{Results on Fault Detection}
\label{sec: fault_detection}

\input{tables/bearings_per_domain_results}
The Bearings dataset~\cite{smith2015rolling} from Case Western Reserve University is widely used for predictive maintenance. It contains 12kHz vibration signals to detect bearings faults in rotating machines. We extract length-4096 samples by a sliding window with stride 290~\cite{zhang2017new}. There are 1 healthy class and 9 fault classes: inner-race fault (IF), outer-race fault (OF), and ball fault (BF) with each subdivided into dimensions 0.007, 0.014 and 0.021 inches. We apply mean shift, scaling and masking augmentations by setting $\mu=\bar{x}$, $\mu_{new}=0$, $\sigma=sd(x)$ and $\sigma_{new}=1$ in Table~\ref{tab:time_series_augmentations}. Samples are augmented with probability 0.5, and additional augmentations are applied with probability 0.5 to allow a mixture of perturbations. There are 8 domains: drive end and fan end location with each operated at 0, 1, 2, and 3 loading torques as in Table~\ref{tab:datasets}. For metadata based similarity, consistency regularization is applied on domains with the same location.
From Table~\ref{tab:bearings_by_domain_results}, the proposed method improves over baseline ERM on almost all target domains. On average, the proposed method attains the best accuracy across all methods with $87.9\%$ and $89.1\%$ given metadata based and learned similarity, respectively. The proposed method also attains lowest calibration errors of $9.2\%$ and $6.2\%$.

\subsection{Results on Human Activity Recognition}
\label{sec: human_activity_recognition}

The HHAR dataset~\cite{stisen2015smart}
consists of multi-channel sensor readings to classify six activities: Biking, Standing, Sitting, Walking, Stair down, and Stair up. Following a recent DA work~\cite{garrett2020sensor}, we focus on smartphone accelerometer readings in the x, y and z direction and extract length-128 samples. Samples are scaled by $\frac{1}{20}$ so that readings for all 3 channels approximately fall between -1 and 1. For this application, mean and standard deviation are known to be important classification features and activities such as Standing are sensitive to abrupt sensor reading changes~\cite{seto2015}, hence we apply limited augmentation i.e. scaling with $\mu=0$, $\sigma=1$ and $\sigma_{new}\sim Unif(0.8,1.2)$. To keep domain number to a suitable level for leave-one-domain-out evaluation, we use 12 domains: the first 3 users each with 4 phone models as in Table~\ref{tab:datasets}. For metadata based similarity, consistency regularization is applied on domains with the same user. 
From Table~\ref{tab:hhar_by_domain_results}, the proposed method improves over the baseline ERM in almost all cases, and has the best accuracy of $88.5\%$ on average. The second-best performing method RSC encourages learning more diverse features by feature masking, and applying the proposed method on RSC by alternating between the two methods further improves average accuracy to $88.9\%$. We chose the alternating procedure~\cite{zhang2021robust} so that strategies from the two methods do not directly interfere with each other. 

We note that although RSC performs similarly to our proposed method, it utilizes masking instead of alignment strategies, and our proposed method significantly improves over all alignment methods evaluated namely Correlation, DANN-DG, CDANN-DG, CORAL-DG and MMD-DG. Furthermore, our proposed method (ECE=$4.1\%, 4.9\%$) learns better calibrated models than RSC (ECE=$5.3\%$).
\input{tables/hhar_per_domain_results}
\input{tables/mimic_per_domain_results}

\subsection{Results on Mortality Prediction}
\label{sec: mortality_prediction}

MIMIC-III~\cite{mimic3data,johnson2016mimic3} is a public clinical dataset for binary in-hospital mortality prediction. We select 11 vital signs that are hourly aggregated, with 24 hours length sample from each patient~\cite{Cai2021TimeSD}. We apply the same data augmentations as in Section~\ref{sec: human_activity_recognition}.
Different age groups represents different domains with Group A: 20-45, Group B: 46-65, Group C: 66-85, Group D: over 85. Due to class imbalance, we downsample the negative class so that both classes have equal sample size. We evaluate performance with AUC following \cite{Cai2021TimeSD}. From Table~\ref{tab:mimic_by_domain_results}, RSC underperforms even ERM on this task, while alignment methods improve performance over baseline ERM. The proposed method has highest AUC ($79.1\%$), and although CORAL-DG achieves similar AUC ($79.0\%$), our proposed method's calibration error is lower by $1.1\%$.

%% file: tables/datasets.tex
\begin{table}
\centering

\caption{Domain attributes. \label{tab:datasets}}

\vspace{-5mm}
\subfloat[Bearings]{
\resizebox{0.33\linewidth}{!}{
\begin{tabular}{*{5}{l}}
\toprule[1pt]\midrule[0.3pt]
\textbf{Loc.}   & \multicolumn{4}{c}{\textbf{Loading torque}}  \\ \cmidrule{2-5}
                & 0 & 1 & 2 & 3 \\ \midrule
Drive           & A & B & C & D \\
Fan             & E & F & G & H \\
\\
\midrule[0.3pt]\bottomrule[1pt]
\end{tabular}}
}
\subfloat[HHAR]{
\resizebox{0.6\linewidth}{!}{
\begin{tabular}{lp{0.9cm}p{0.9cm}p{0.9cm}p{0.9cm}}
\toprule[1pt]\midrule[0.3pt]
\textbf{User}   & \multicolumn{4}{c}{\textbf{Phone model}}  \\ \cmidrule{2-5}
                    & Nexus    & S3  & S3 mini   & S+ \\ \midrule
User 1              & A & B & C & D \\
User 2              & E & F & G & H \\
User 3              & I & J & K & L \\
\midrule[0.3pt]\bottomrule[1pt]
\end{tabular}}
}

\vspace{-5mm}
\end{table}

%% file: tables/bearings_per_domain_results.tex
\begin{table*}[ht]
\centering

\caption{Bearings: Classification accuracy and expected calibration error (ECE) on target domain using leave-one-domain-out testing. 
\label{tab:bearings_by_domain_results}}
\vspace{-3mm}

\resizebox{0.7\linewidth}{!}{
\begin{tabular}{*{10}{l}c}
\toprule[1pt]\midrule[0.3pt]
\textbf{Method}     & \multicolumn{9}{c}{\textbf{Accuracy ($\%$) $\uparrow$}}  & \multicolumn{1}{c}{\textbf{ECE ($\%$) $\downarrow$}}  \\ \cmidrule(lr){2-10} \cmidrule(lr){11-11}
                    & A     & B     & C     & D     & E     & F     & G     & H     & Avg  & Avg \\ \midrule
ERM                 & 65.4  & 93.8  & 96.0  & 71.2  & 68.4  & 83.0  & 94.0  & 86.2  & 82.2 & 14.2\\
IRM                 & 60.7  & 87.0  & 89.2  & 76.0  & 62.8  & 80.9  & 92.8  & 88.5  & 79.7 & 15.9\\
GroupDRO            & 55.7  & 70.0  & 77.8  & 74.8  & 60.7  & 59.2  & 65.6  & 50.5  & 64.3 & 31.9\\
Interdomain Mixup   & 62.0  & 86.5  & 96.8  & 76.0  & 82.0  & 95.4  & 97.7  & 87.2  & 85.4 & 13.8\\
MLDG                & 62.8  & 77.6  & 85.9  & 72.8  & 63.3  & 58.5  & 60.7  & 55.0  & 67.1 & 30.8\\
MTL                 & 35.3  & 64.5  & 66.6  & 48.3  & 47.6  & 48.2  & 36.7  & 44.9  & 49.0 & 44.4\\
Correlation         & 46.5  & 79.0  & 90.0  & 69.1  & 71.5  & 85.5  & 80.5  & 83.9  & 75.7 & 18.2\\
VREx                & 63.8  & 90.5  & 97.2  & 81.6  & 70.3  & 83.9  & 92.0  & 84.3  & 83.0 & 14.3\\
RSC                 & 62.4  & 94.4  & 98.0  & 86.5  & 73.4  & 87.4  & 97.3  & 85.7  & 85.6 & 12.0\\
DANN-DG             & 56.2  & 84.7  & 92.2  & 80.2  & 70.0  & 79.1  & 89.1  & 90.5  & 80.3 & 15.1\\
CDANN-DG            & 56.0  & 80.8  & 94.8  & 80.2  & 70.7  & 81.4  & 90.3  & 84.0  & 79.8 & 15.7\\ 
CORAL-DG            & 62.5  & 77.9  & 90.0  & 76.0  & 63.1  & 79.2  & 74.5  & 83.0  & 75.8 & 19.7\\
MMD-DG              & 53.9  & 67.7  & 84.2  & 67.6  & 63.2  & 74.3  & 74.7  & 56.7  & 67.8 & 28.7\\ \hdashline
Ours (Metadata sim.) 
                    & 86.8  & 95.3  & 97.6  & 79.8  & 77.4  & 82.7  & 93.4  & 90.8  & \underline{87.9} & \underline{9.2}\\
Ours (Learned sim.)
                    & 89.1  & 97.9  & 97.1  & 75.8  & 81.5  & 85.3  & 94.4  & 91.8  & \textbf{89.1} & \textbf{6.2}\\
\midrule[0.3pt]\bottomrule[1pt]
\end{tabular}
}

\vspace{-5mm}
\end{table*}

%% file: tables/hhar_per_domain_results.tex
\begin{table*}
\centering

\caption{HHAR: Classification accuracy and expected calibration error (ECE) on target domain using leave-one-domain-out testing. 
\label{tab:hhar_by_domain_results}}
\vspace{-3mm}

\resizebox{0.9\linewidth}{!}{
\begin{tabular}{*{14}{l}c}
\toprule[1pt]\midrule[0.3pt]
\textbf{Method}     & \multicolumn{13}{c}{\textbf{Accuracy ($\%$) $\uparrow$}} & \multicolumn{1}{c}{\textbf{ECE ($\%$) $\downarrow$}} \\ \cmidrule(lr){2-14} \cmidrule(lr){15-15}
        & A     & B     & C     & D     & E     & F     & G     & H     & I     & J     & K     & L     & Avg  & Avg \\ \midrule
ERM     & 86.4  & 91.6  & 81.0  & 91.7  & 71.3  & 96.9  & 96.4  & 85.9  & 85.0  & 88.1  & 86.6  & 89.5  & 87.5 & 6.0\\
IRM     & 87.2  & 92.0  & 80.0  & 90.5  & 71.7  & 96.5  & 96.6  & 85.3  & 84.3  & 88.4  & 87.3  & 89.6  & 87.4 & 6.2\\
GroupDRO& 80.4  & 76.7  & 52.4  & 74.6  & 63.6  & 77.3  & 76.6  & 75.0  & 86.3  & 86.8  & 82.8  & 70.3  & 75.2 & 15.2\\
Interdomain Mixup   
        & 80.2  & 68.9  & 61.4  & 69.7  & 55.3  & 71.1  & 81.4  & 64.7  & 87.8  & 85.0  & 84.9  & 71.5  & 73.5 & 13.5\\
MLDG    & 81.7  & 75.8  & 58.8  & 79.3  & 58.4  & 70.7  & 70.9  & 68.0  & 86.8  & 87.7  & 84.7  & 70.4  & 74.4 & 18.2\\
MTL     & 79.6  & 75.8  & 60.9  & 77.1  & 62.8  & 75.7  & 80.2  & 72.7  & 85.7  & 81.1  & 79.3  & 71.9  & 75.2 & 17.9\\
Correlation
        & 80.3  & 91.0  & 81.7  & 87.9  & 69.1  & 95.4  & 95.8  & 88.5  & 85.1  & 85.6  & 88.6  & 88.9  & 86.5 & 5.8\\
VREx    & 87.1  & 90.6  & 80.5  & 92.2  & 71.0  & 96.5  & 96.7  & 85.5  & 85.5  & 88.7  & 87.5  & 90.2  & 87.7 & 5.7\\
RSC     & 87.3  & 90.5  & 84.4  & 92.2  & 73.9  & 96.7  & 96.9  & 86.2  & 86.8  & 87.5  & 88.5  & 90.1  & 88.4 & 5.3\\
DANN-DG & 84.7  & 89.5  & 72.4  & 92.8  & 71.2  & 95.1  & 94.8  & 84.2  & 81.6  & 84.3  & 84.9  & 86.7  & 85.2 & 7.5\\
CDANN-DG& 85.6  & 86.0  & 79.8  & 89.6  & 72.4  & 93.6  & 95.6  & 83.0  & 81.3  & 87.0  & 83.4  & 85.8  & 85.2 & 8.2\\ 
CORAL-DG& 80.5  & 76.8  & 58.4  & 74.3  & 62.5  & 77.5  & 85.8  & 74.2  & 86.8  & 79.7  & 86.2  & 69.1  & 76.0 & 15.4\\
MMD-DG  & 81.9  & 74.0  & 52.9  & 75.4  & 60.3  & 76.8  & 79.0  & 74.6  & 86.9  & 85.6  & 83.7  & 68.9  & 75.0 & 17.2\\  \hdashline
Ours (Metadata sim.) 
        & 87.4  & 91.0  & 80.7  & 94.6  & 75.7  & 96.5  & 97.1  & 86.2  & 85.0  & 89.2  & 89.1  & 89.8  & \underline{88.5} & \textbf{4.1}\\
Ours (Learned sim.) 
        & 87.3  & 90.6  & 85.3  & 93.5  & 76.0  & 96.1  & 96.7  & 86.0  & 85.1  & 88.1  & 88.6  & 88.9  & \underline{88.5} & 4.9\\ \hdashline
RSC \\
+ Ours (Metadata sim.) 
        & 87.2  & 89.9  & 86.0  & 93.6  & 75.6  & 96.6  & 96.4  & 86.1  & 85.7  & 88.6  & 90.4  & 90.6  & \textbf{88.9} & \underline{4.5}\\
+ Ours (Learned sim.)
        & 86.3  & 89.0  & 84.3  & 93.3  & 74.4  & 96.8  & 96.9  & 86.8  & 87.0  & 86.7  & 90.4  & 89.2  & \underline{88.5} & 4.8\\
\midrule[0.3pt]\bottomrule[1pt]
\end{tabular}
}

\vspace{-5mm}
\end{table*}

%% file: tables/mimic_per_domain_results.tex
\begin{table}[ht]
\centering

\caption{MIMIC-III: AUC and expected calibration error (ECE).
\label{tab:mimic_by_domain_results}}
\vspace{-3mm}

\resizebox{0.95\linewidth}{!}{
\begin{tabular}{*{6}{l}c}
\toprule[1pt]\midrule[0.3pt]
\textbf{Method}     & \multicolumn{5}{c}{\textbf{AUC ($\%$) $\uparrow$}}  & \multicolumn{1}{c}{\textbf{ECE ($\%$) $\downarrow$}}  \\ \cmidrule(lr){2-6} \cmidrule(lr){7-7}
                    & A     & B     & C     & D     & Avg  & Avg \\ \midrule
ERM                 & 84.9  & 78.0  & 77.4  & 73.1  & 78.4 & 7.6\\
RSC                 & 83.5  & 78.1  & 77.5  & 73.5  & 78.2 & 8.5\\
DANN-DG             & 84.9  & 79.5  & 76.2  & 74.0  & 78.7 & \underline{7.4}\\
CDANN-DG            & 85.2  & 79.2  & 76.1  & 75.1  & 78.9 & 8.2\\
CORAL-DG            & 86.3  & 78.0  & 77.1  & 74.7  & \underline{79.0} & 7.5\\
MMD-DG              & 85.5  & 77.8  & 77.0  & 73.6  & 78.5 & 8.2\\  \hdashline
Ours (Learned sim.)
                    & 86.5  & 78.1  & 77.5  & 74.4  & \textbf{79.1} & \textbf{6.4}\\
\midrule[0.3pt]\bottomrule[1pt]
\end{tabular}
}

\vspace{-4mm}
\end{table}

%% file: sections/further_analysis.tex
\section{Further Analysis}
\label{sec: further_analysis}

\input{tables/ablation}
\input{tables/bearings_diff_dist} 
\input{tables/different_clustering}
\textbf{Ablation study:} 
In Table~\ref{tab:ablation_all}, we see that applying data augmentations and consistency regularization individually improves model performance over ERM for both datasets. Performance increases further when both strategies are applied. 

\textbf{Effect of design choices for regularizer:} 
To further study the regularizer in isolation, we apply the proposed method with metadata based selection and no time series augmentations. 

\emph{Distance functions:} We experiment with 3 distance functions: squared $L_2$ distance, cosine distance and KL-divergence, either between individual samples and cluster centroids or between domain and cluster centroids. Regularization is applied on the features $\boldsymbol{z}$, logits $\boldsymbol{g}$ or soft labels $\boldsymbol{s}$. Comparing the generalization performance in Table~\ref{tab:bearings_diff_dist}, domain-level regularization tends to have higher accuracy, possibly because it allows greater diversity of representations in each domain. 

\emph{Logits vs Soft labels vs Features:} Regularizing on logits results in higher accuracy in most cases (Table~\ref{tab:bearings_diff_dist}). It allows more flexibility as both feature extractor and classifier are directly regularized while preserving class-relationships. Soft labels are normalized logits and have limited variability across source domains for further alignment. The feature space is generally much larger than the logit or label space, and hence possibly more difficult to effective align. We observe that all choices of distance functions and representations attain better performance than ERM ($82.2\%$), with regularization of logits at the domain-level achieving the best accuracy of $87.1\%$.

\emph{Cluster assignments:} Next, we study the effect of cluster assignments on DG performance for each target domain in the Bearings dataset in Table~\ref{tab:diff_clustering}. Consistency regularization tends to improve performance over ERM across all target domains. While regularization with 4 clusters and 1 cluster both improve accuracy over ERM by $2.2\%$ and $3.4\%$ respectively, regularization with 2 clusters achieves the highest improvement of $4.9\%$. Domains in each of the two clusters have the same machine location, and hence can be expected to be closely-related with similar class relationships. This shows good cluster assignments are critical for good performance. 

\begin{figure}[ht!]
    \centering
    
    \vspace{-5mm}
    \subfloat[Loss components\label{fig:optimization_loss}]{\includegraphics[width=0.16\textwidth]{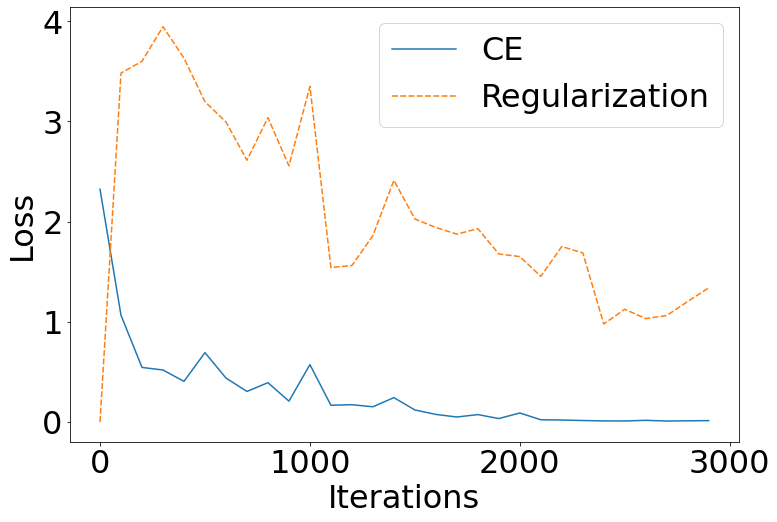}}
    \subfloat[Training loss\label{fig:optimization_cluster}]{\includegraphics[width=0.16\textwidth]{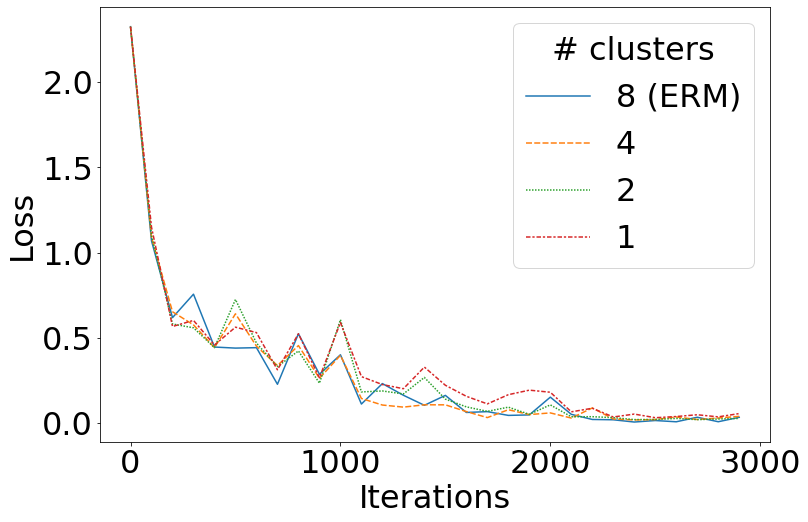}}
    \subfloat[Training time\label{fig:optimization_time}]{\includegraphics[width=0.16\textwidth]{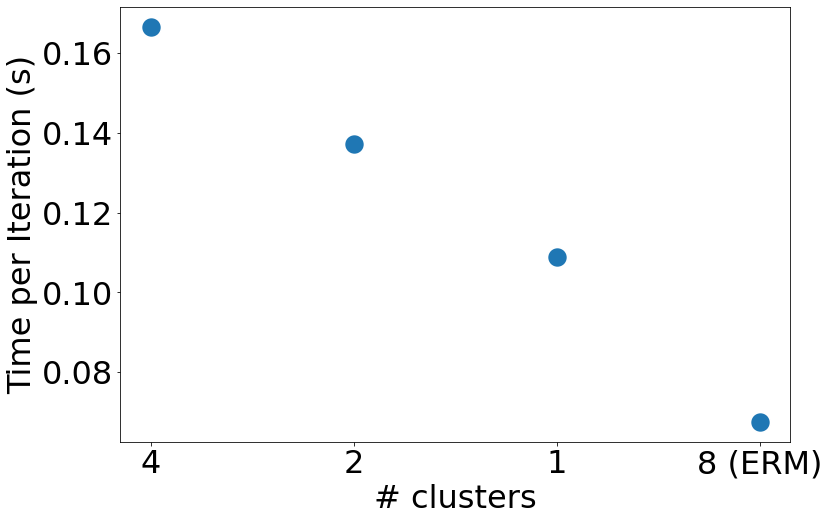}}    
    \vspace{-1mm}     
    \caption{Bearings: Loss and average computation time with target H. ERM is equivalent to no regularization or assigning each domain to its own cluster. \label{fig:optimization}}
    \vspace{-3mm}
\end{figure}

\emph{Training process:} 
Figure~\ref{fig:optimization_loss} shows loss components using regularization with 2 clusters, where the model initially prioritized learning the classification task and then increased domain alignment as training progresses.
Figure~\ref{fig:optimization_cluster} and \ref{fig:optimization_time} show loss and computation time per iteration (total 3000 iterations) for different number of clusters ran on Tesla V100-SXM2 GPU. Hyperparameter is fixed at $\lambda=0.01$. Training loss converges and computation time is low for all cluster numbers tested. 

\textbf{Analysis of target risk bound:}
In multi-source DA, the target domain risk is bounded by a sum consisting empirical source domain risks and divergence between source and target domains, as stated in Theorem~\ref{theorem: DA}.

\begin{theorem}~\cite{Zhao2018AdversarialMS}
Let $\mathcal{H}$ be a hypothesis class with VC dimension $v$. For source domain $\{\hat{S}_m\}_{m=1}^M$ with $\frac{N}{M}$ i.i.d. samples from each domain, and target domain $\hat{T}$ with $N$ samples, with probability $1-\delta$, for all $\gamma \in \mathcal{H}$ and $\forall a \in \mathbb{R}^M_+$ such that $\sum_{m=1}^M a_m = 1$,
\begin{align}
    \epsilon_T(\gamma) \leq 
    & \sum_{m=1}^M a_i \left( \hat{\epsilon}_{S_m}(\gamma) + \frac{1}{2}div_{\mathcal{H}}(\hat{T}; {\hat{S}_m}) \right) +
    r_a \nonumber \\ 
    & + O\left( \sqrt{ \frac{1}{N} \left( \log\frac{1}{\delta} + v\log\frac{N}{v} \right) } \right) \label{eqn: task_risk_bound}
\end{align}
where 
$\epsilon(\gamma)$ and $\hat{\epsilon}(\gamma)$ are risk and empirical risk of $\gamma$ w.r.t. the true labeling function respectively,
$r_a$ is upper bound of $\text{inf}_{\gamma \in \mathcal{H}} \left( \sum_{m=1}^M a_m \epsilon_{S_m}(\gamma) + \epsilon_{T}(\gamma) \right)$, and $\mathcal{H}$-divergence $div_\mathcal{H}(T; S) \defeq 2\text{sup}_{A \in \mathcal{A}_\mathcal{H}} \| P_T(A) - P_S(A)\|$ with $A_\mathcal{H} \defeq \{\gamma^{-1}(\{1\}) | \gamma \in \mathcal{H}\}$. \label{theorem: DA}
\end{theorem}

In DA, adversarial learning and regularization are used to align source and target domains to directly reduce $\mathcal{H}$-divergence.
Unlike DA, target domain $\hat{T}$ is unavailable during training for DG. Aligning source domains helps to prevent learning spurious features, but aligning all source domains as in existing domain alignment methods does not guarantee reducing $\mathcal{H}$-divergence.

We create a two-channel toy dataset for binary classification of sawtooth and square-shaped signals to compare the bound terms and empirical target risk for different methods; we leave out the $r_a$ term since we use the same backbone network for all methods.
Domain A and B have shaped signals on channel 1. Domain C and D additionally have square signals on channel 2 for the square signal class. Noise added to signal is Gaussian distributed with standard deviation 2 for A and B and 4 for C and D. We use D as target and the rest as source.

We implement our proposed method with regularization on domain A and B since C differs by having additional channel 2 features.
In Table~\ref{tab:target_risk}, lower $\mathcal{H}$-divergence indicates lower target risk. Our proposed method has lower $\mathcal{H}$-divergence and target risk than DANN-DG and CDANN-DG. By not aligning domain C, we also allow the model to retain channel 2 features which target domain D can use for classification. More details are in Appendix B.
\input{tables/target_risk}

%% file: tables/ablation.tex
\begin{table}[ht]
\centering
\caption{Effect of regularization and augmentation strategies. \label{tab:ablation_all}}
\vspace{-5mm}

\subfloat[Regularization and augmentation]{
\resizebox{0.55\linewidth}{!}{
\begin{tabular}{cccc}
\toprule[1pt]\midrule[0.3pt]
\multicolumn{2}{c}{\textbf{Strategy}}                  
& \multicolumn{2}{c}{\textbf{Avg Acc ($\%$)}} \\ \cmidrule{1-4}
Reg             & Aug           & Bearings  & HHAR \\ \midrule
\xmark          & \xmark        & 82.2      & 87.5 \\
\xmark          & \cmark        & 86.5      & 88.1 \\ \hdashline
\multicolumn{1}{c}{(Metadata sim.)} & \multicolumn{1}{c}{} \\
\cmark          & \xmark        & 87.1      & 88.5 \\
\cmark          & \cmark        & 87.9      & 88.5 \\ \hdashline
\multicolumn{1}{c}{(Learned sim.)} & \multicolumn{1}{c}{} \\
\cmark          & \xmark        & 86.8      & 88.3 \\
\cmark          & \cmark        & 89.1      & 88.5 \\
\midrule[0.3pt]\bottomrule[1pt]
\end{tabular}
}}
\hspace{0.05cm}
\subfloat[Bearings: Time series augmentation without consistency regularization]{
\resizebox{0.35\linewidth}{!}{
\begin{tabular}{lc}
\toprule[1pt]\midrule[0.3pt]
\multicolumn{1}{c}{\textbf{Aug}} 
& \multicolumn{1}{c}{\textbf{Avg Acc ($\%$)}} \\ \midrule
None            & 82.2 \\
Mean shift      & 83.0 \\
Scale           & 82.4 \\
Mask            & 82.4 \\
All             & 86.5 \\
\midrule[0.3pt]\bottomrule[1pt]
\end{tabular}
}}

\vspace{-5mm}
\end{table}

%% file: tables/bearings_diff_dist.tex
\begin{table}[ht]
\centering

\caption{Bearings: Accuracy using different regularization functions. \label{tab:bearings_diff_dist}}
\vspace{-3mm}

\resizebox{0.8\linewidth}{!}{
\begin{tabular}{lllllll}
\toprule[1pt]\midrule[0.3pt]
& \multicolumn{6}{c}{\textbf{Avg Accuracy ($\%$)}} \\ \cmidrule{2-7}
& \multicolumn{3}{c}{\textit{Sample-level}}   & \multicolumn{3}{c}{\textit{Domain-level}}  \\ \cmidrule{2-7}
Regularize on               & L2    & cos   & KL    & L2    & cos   & KL\\ \midrule
Features $\boldsymbol{z}$   & 83.9  & 84.9  & N/A   & 85.4  & 82.3  & N/A\\
Logits $\boldsymbol{g}$     & 86.2  & 84.2  & N/A   & \textbf{87.1} & 85.9 & N/A\\
Soft labels $\boldsymbol{s}$& 82.9  & 83.3  & 82.8  & 82.5  & 83.8  & 83.8\\
\midrule[0.3pt]\bottomrule[1pt]
\end{tabular}}

\vspace{-3mm}
\end{table}

%% file: tables/different_clustering.tex
\begin{table}[ht!]
\centering

\caption{Bearings: Accuracy given different cluster assignments. \label{tab:diff_clustering}}
\vspace{-3mm}

\resizebox{0.85\linewidth}{!}{
\begin{tabular}{*{3}{c}}
\toprule[1pt]\midrule[0.3pt]
\textbf{Cluster assignment}    & \textbf{\# clusters}      & \textbf{Avg Acc ($\%$)} \\ \midrule
\{A\},\{B\},\{C\},\{D\},\{E\},\{F\},\{G\},\{H\} & 8 (ERM) & 82.2    \\
\{A,B\},\{C,D\},\{E,F\},\{G,H\} & 4 & 84.4  \\
\{A,B,C,D\},\{E,F,G,H\} & 2 & \textbf{87.1}    \\
\{A,B,C,D,E,F,G,H\} & 1 & \underline{85.6} \\
\midrule[0.3pt]\bottomrule[1pt]
\end{tabular}}

\vspace{-5mm}
\end{table}

%% file: tables/target_risk.tex
\begin{table}[ht]
\centering
\vspace{-3mm}

\caption{Toy Dataset: Comparison of bound terms and target risk. \label{tab:target_risk}}
\vspace{-3mm}

\resizebox{0.7\linewidth}{!}{
\begin{tabular}{l*{3}{c}}
\toprule[1pt]\midrule[0.3pt]
\textbf{Method} & \textbf{Source Risk}  & \textbf{$\mathcal{H}$-divergence}   & \textbf{Target Risk} \\ \midrule
DANN-DG    & 0.196 & 1.597 & 0.399 \\
CDANN-DG   & 0.198 & 1.591 & 0.333 \\
Ours    & 0.196 & 1.536 & 0.111 \\
\midrule[0.3pt]\bottomrule[1pt]
\end{tabular}}

\vspace{-3mm}
\end{table}

%% file: sections/conclusion.tex
\section{Conclusion}
\label{sec: conclusion}

In this work, we introduced a representation learning method for domain generalization for time series classification. We applied augmentations to improve sample diversity, and selective consistency regularization to enforce similar predictions for similar domains.
From comprehensive experiments, we showed that the proposed method significantly improves over ERM and performs competitively compared to state-of-the-art methods in both classification accuracy and model calibration. 

%% file: sections/appendix_implementation_details.tex
\subsection{Implementation Details}
\label{sec: implementation_details}

We provide additional details on the implementation for our experiments and the datasets used.

We evaluate the proposed method on three real-world time series datasets for fault detection, human activity recognition and mortality prediction following the evaluation procedure in \cite{gulrajani2020domainbed}.
For each dataset, we use leave-one-domain-out evaluation where we treat each domain as the unseen target in turn and train with rest as source domains. Each source-target combination is evaluated over 3 random seeds, and each method is tuned with 20 random hyperparameter configurations with the best configuration selected using validation accuracy. All methods use the same backbone networks specific to each dataset.

\subsubsection{Hyperparameters}

We fix learning rate $0.001$, weight decay $5\times 10^{-5}$, and batch size $32$ per domain. Models are trained for $3000$ iterations, with learning rate reduced by a factor of $10$ after 2400 iterations. All other hyperparameters are tuned by random sampling from distributions in Table~\ref{tab:hyperparameter}. All experiments are run using Adam optimizer with the NVIDIA container image for PyTorch, release 20.03.

\subsubsection{Datasets and Network Architectures}

We provide details on the sample size of the datasets. Backbone network architectures used for each dataset is given in Table~\ref{tab:networks}.

\textbf{Bearings:}
All domains have the same number of samples. For each domain, the sample size of each class is `normal': 416, `IF:0.007': 371, `BF:0.007': 409, `OF:0.007': 417, `IF:0.014': 387, `BF:0.014': 408, `OF:0.014': 398, `IF:0.021': 407, `BF:0.021': 383, `OF:0.021': 404. We use a 6-layer CNN as feature extractor and a 3-layer FCN as classifier~\cite{zhang2021robust}.

\textbf{HHAR:} 
Sample size differs across domain according to availability of data per user and device, as in Table~\ref{tab:datasets_samplesize}.
We use a 3-layer CNN as feature extractor and a 1-layer fully-connected network as classifier~\cite{liu2016adaptive}.

\textbf{MIMIC-III:}
All classes are sampled to be of the same size. The sample size of each class for the 4 domains are A: 164, B: 625, C: 1179, D: 495. We use the same network architecture as for HHAR but with smaller hidden dimension size.

%% file: sections/appendix_further_experiment_results.tex
\subsection{Further Experiment Results}
\label{sec: further_experiment_results}

\begin{figure}[ht]
    \centering
    \vspace{-8mm}
        
    \subfloat[Bearings]{\includegraphics[width=0.2\textwidth]{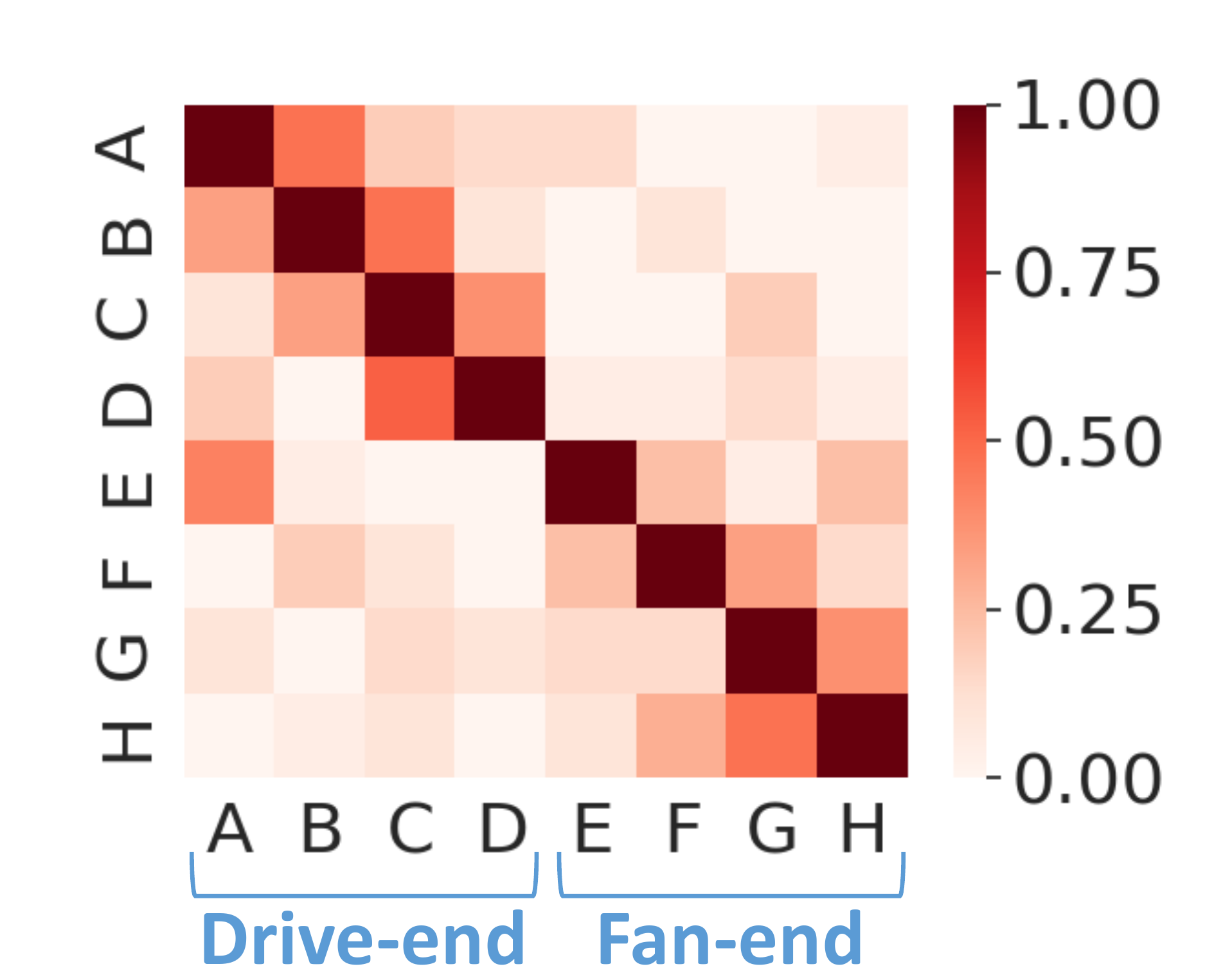}}
    \subfloat[HHAR]{\includegraphics[width=0.2\textwidth]{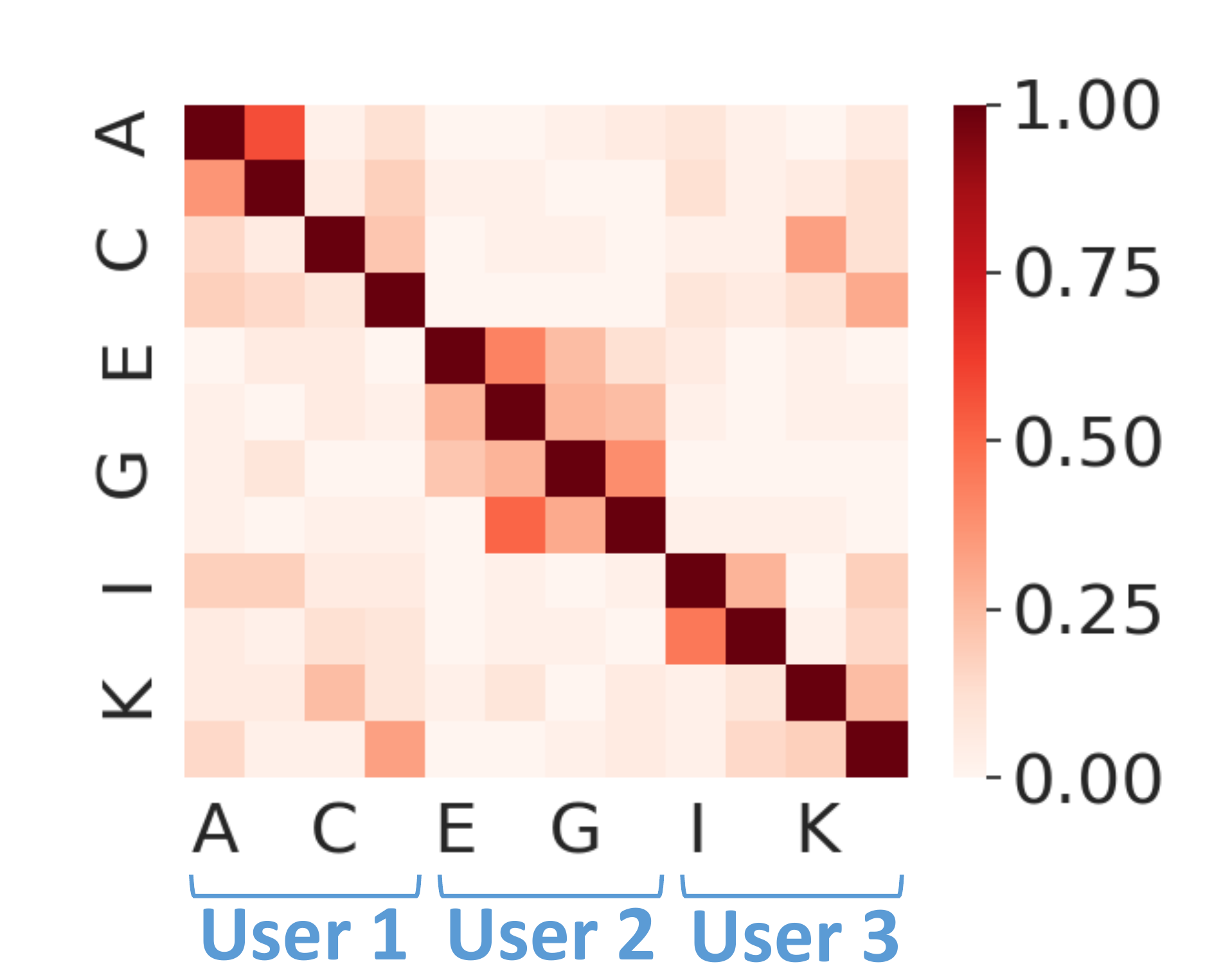}}
    \vspace{-1mm}     
    \caption{Fraction of runs where domain $j$ (column) is the nearest neighbor of domain $i$ (row) at end of training. Diagonal values are 1. \label{fig:cluster}}
    
\end{figure}

\textbf{Visualization of learned domain relationships on Bearings and HHAR:}
The proposed method estimates inter-domain relationships when domain metadata are not provided. Figure~\ref{fig:cluster} plots the proportion of runs each pair of domain is estimated to be closest neighbors. For Bearings, the estimated clusters approximately match location groupings. For HHAR, the variation between phone models (i.e. Nexus and S3 versus S3 mini and S+) appears larger than that between some users (i.e. User 1 versus User 3). Domain relationships estimated from data can be different from those inferred from metadata descriptions and can contain finer measures of inter-domain similarity. In fact, using learned similarity obtains higher accuracy (89.1\%) than using fixed metadata-inferred similarity (87.9\%) for Bearings.

\textbf{Additional experiments on Bearings and HHAR:}
We additionally evaluate on a more challenging setting where target domain conditions are not combinations of source domain conditions. For Bearings, we use only domains from drive-end location, so each domain has a distinct loading torque. For HHAR, we use only domains from the first user, so each domain has a distinct phone model. We fix the hyperparameter of our proposed method with learned inter-domain similarity to $\lambda=0.01$ and $\xi=0.1$. From Table~\ref{tab:bearings_locdrive} and \ref{tab:hhar_user1}, our proposed method outperforms ERM on all target domains in Bearings, and on average in HHAR. 
\input{tables/bearings_locdrive}
\input{tables/hhar_user1}

\input{tables/ablation_learned_reg}

\input{tables/kernel_choice}

\begin{figure*}[ht]
    \centering
        
    \subfloat{\includegraphics[width=0.2\textwidth]{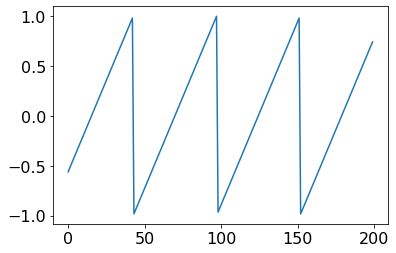}}
    \subfloat{\includegraphics[width=0.2\textwidth]{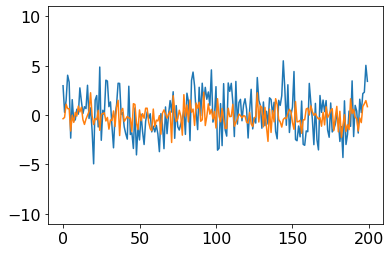}}
    \subfloat{\includegraphics[width=0.2\textwidth]{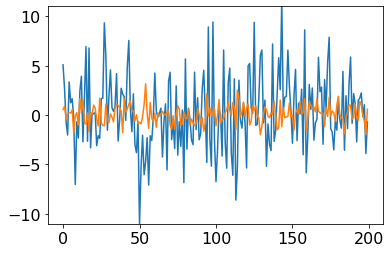}}
    \subfloat{\includegraphics[width=0.2\textwidth]{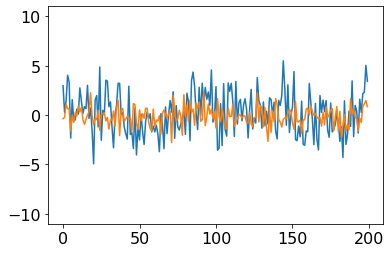}}
    \subfloat{\includegraphics[width=0.2\textwidth]{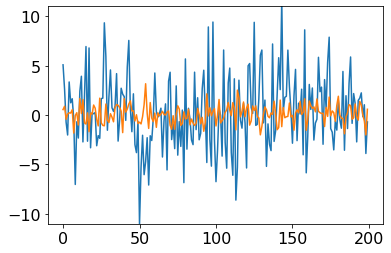}}
    
    \setcounter{subfigure}{0}
    \subfloat[Signal w/o noise]{\includegraphics[width=0.2\textwidth]{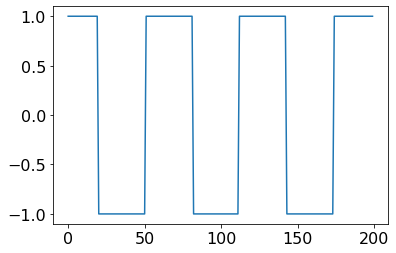}}
    \subfloat[Domain A]{\includegraphics[width=0.2\textwidth]{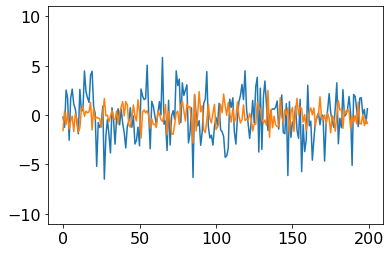}}
    \subfloat[Domain B]{\includegraphics[width=0.2\textwidth]{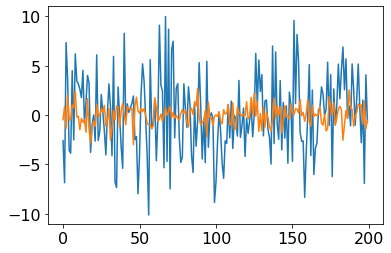}}
    \subfloat[Domain C]{\includegraphics[width=0.2\textwidth]{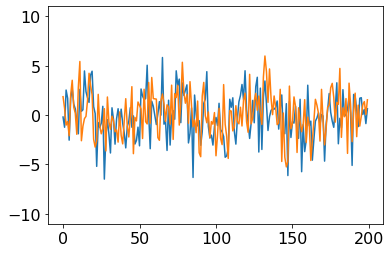}}
    \subfloat[Domain D]{\includegraphics[width=0.2\textwidth]{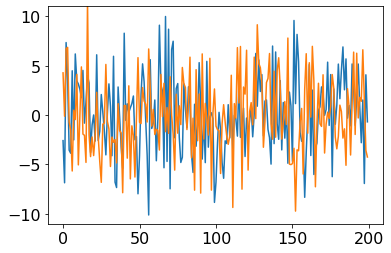}}
    \vspace{-1mm}     
    \caption{Toy dataset: Example series. (a) Clean sawtooth and square signals used for data generation, (b)-(e) Samples generated for domain A to D, with channel 1 in blue and channel 2 in orange. Best viewed in color. \label{fig:toy_dataset}}
    
\end{figure*}

\input{tables/toy_dataset}
\input{tables/target_risk_full}

\input{tables/hyperparameter}
\input{tables/networks}
\input{tables/datasets_samplesize}

\textbf{Ablation study:}
We study the effect of neighbor domain selection and weight function in regularization with learned inter-domain similarity on Bearings. Hyperparameters are fixed at $\lambda=0.01$ and $\xi=0.1$. In Table~\ref{tab:ablation_learned_reg}, regularizing each domain with its nearest neighbor achieves higher accuracy than regularizing random pairs of domains. While nearest neighbor selection and a fixed weight of 1 has best accuracy in this study, we note that the RBF hyperparameter $\xi$ can be tuned and the RBF weight approaches 1 as $\xi\rightarrow 0$. 

Lastly, we study the use of different kernel functions in MMD for distribution matching in our proposed method. The linear kernel is the default used in all our experiments, and the Gaussian kernel is implemented as in \cite{domainbedgit}. We do not apply time series augmentations in order to study the effect of regularization in isolation. From Table~\ref{tab:kernel_choice}, the linear kernel has better performance than the Gaussian kernel for both Bearings and HHAR datasets.

\textbf{Analysis of target risk bound:}
We create a toy dataset according to domain attributes in Table~\ref{tab:toy_attributes} for binary classification of sawtooth and square-shaped signals to compare the target risk bound terms in main manuscript Theorem 1 and empirical target risk. Each sample is a length-200 time series with two channels. Domain A and B only have sawtooth and square signals on channel 1. Domain C and D additionally have square signals on channel 2 for the square signal class. Signals are generated with number of cycles uniformly at random in $[\frac{1}{6}, 5]$, and noise added to the signals is Gaussian distributed with standard deviation 2 for domain A and B and 4 for domain C and D. We add Gaussian $N(0,1)$ noise to channels with no shape signals. We illustrate example series from each domain in Figure~\ref{fig:toy_dataset}. We use domain D as unseen target domain and the rest as source domains. Each source domain has 400 samples and target domain has 1200 samples following sample size specifications in main manuscript Equation 1.

We provide estimates of the target risk bound terms in Table~\ref{tab:target_risk_full} on models trained using existing domain alignment methods (i.e. DANN-DG, CDANN-DG) and our proposed method with consistency regularization applied on domain A and B. We do not regularize domain C in our method because it is the most dissimilar from the other two source domains by having additional channel 2 features that are correlated with signal class. All regularization hyperparameters are set to 1 i.e. $\lambda=1$, and we use the same network architecture as for HHAR with 1-layer CNN. We do not apply time series augmentations in order to study the effect of regularization in isolation. To control for possible differences due to model convergence, we stop training for all methods when average source risk falls below 0.2. We follow the procedure in \cite{Ganin2016DANN} to estimate $\mathcal{H}$-divergence. For each source $S_m$ and target $T$ domain pair:
\begin{enumerate}
    \item Construct domain classification training set with 200 samples from source domain and 200 samples randomly selected from target domain, and construct domain classification test set with the other 200 samples from source domain and 200 samples randomly selected from the remainder of target domain; 
    \item Train SVMs with RBF kernel for binary domain classification with regularization hyperparameter $C \in \{0.001, 0.01, 0.1, 1, 10, 10, 100\}$;
    \item Select $C$ with lowest training error, and evaluate selected SVM on test set to obtain test error $\rho$;
    \item Approximate the $\mathcal{H}$-divergence by $div_\mathcal{H}(\hat{S}_m, \hat{T}) = 2(1-2\rho)$.
\end{enumerate}

We observe from Table~\ref{tab:target_risk_full} that a lower $\mathcal{H}$-divergence tend to indicate a lower target risk. Our proposed method has the lowest $\mathcal{H}$-divergence, especially between domain B and D. Source domain B is aligned with only one domain in our method instead of two in other methods, and it is the most similar to target domain D in terms of signal noise level.
By not regularizing source domain C, we also allow the model to retain channel 2 signal features for classification on target domain D.

%% file: tables/bearings_locdrive.tex
\begin{table}[htb]
\centering
\vspace{-3mm}

\caption{Bearings drive-end domains: Classification accuracy and expected calibration error (ECE) on target domain. \label{tab:bearings_locdrive}}

\resizebox{\linewidth}{!}{
\begin{tabular}{*{6}{l}c}
\toprule[1pt]\midrule[0.3pt]
\textbf{Method}     & \multicolumn{5}{c}{\textbf{Accuracy ($\%$) $\uparrow$}}  & \textbf{ECE ($\%$) $\downarrow$} \\ \cmidrule(lr){2-6} \cmidrule(lr){7-7}
                    & A     & B     & C     & D     & Avg   & Avg \\ \midrule
ERM                 & 84.4  & 94.9  & 98.9  & 86.0  & 91.0  & 5.2 \\
Ours (Learned sim.)
                    & \textbf{93.9}  & \textbf{97.8}  & \textbf{99.0}  & \textbf{88.7}  & \textbf{94.8} & \textbf{3.1} \\
\midrule[0.3pt]\bottomrule[1pt]
\end{tabular}
}

\end{table}

%% file: tables/hhar_user1.tex
\begin{table}[htb]
\centering
\vspace{-3mm}

\caption{HHAR user 1: Classification accuracy and expected calibration error (ECE) on target domain. \label{tab:hhar_user1}}

\resizebox{\linewidth}{!}{
\begin{tabular}{*{6}{l}c}
\toprule[1pt]\midrule[0.3pt]
\textbf{Method}     & \multicolumn{5}{c}{\textbf{Accuracy ($\%$) $\uparrow$}} & \textbf{ECE ($\%$) $\downarrow$}\\ \cmidrule(lr){2-6} \cmidrule(lr){7-7}
        & A     & B     & C     & D     & Avg   & Avg \\ \midrule
ERM     & \textbf{79.0}  & 80.5  & \textbf{71.5}  & 80.3  & 77.8    & 14.6 \\
Ours (Learned sim.) 
        & 78.7  & \textbf{81.8}  & 70.8  & \textbf{82.0}  & \textbf{78.3}   & \textbf{13.4} \\
\midrule[0.3pt]\bottomrule[1pt]
\end{tabular}
}

\end{table}

%% file: tables/ablation_learned_reg.tex
\begin{table}[ht]
\centering
\vspace{-3mm}

\caption{Bearings: Regularization strategies for regularization with learned inter-domain similarity. \label{tab:ablation_learned_reg}}

\begin{tabular}{ccc}
\toprule[1pt]\midrule[0.3pt]
\textbf{Selected neighbor} & \textbf{Weight function} & \textbf{Avg Accuracy ($\%$)} \\ \midrule
Random  & 1                 & 80.9 \\
Random  & RBF ($\xi=0.1$)   & 81.2 \\
Nearest & 1                 & \textbf{84.9} \\
Nearest & RBF ($\xi=0.1$)   & \underline{83.2} \\
\midrule[0.3pt]\bottomrule[1pt]
\end{tabular}

\end{table}

%% file: tables/kernel_choice.tex
\begin{table}[h]
\centering
\vspace{-3mm}

\caption{Choice of MMD kernel function. \label{tab:kernel_choice}}

\resizebox{0.6\linewidth}{!}{
\begin{tabular}{lcc}
\toprule[1pt]\midrule[0.3pt]
\textbf{Kernel}                  
& \multicolumn{2}{c}{\textbf{Avg Acc ($\%$)}} \\ \cmidrule{2-3}
            & Bearings  & HHAR \\ \midrule
(Metadata sim.) \\
Gaussian    & 83.1      & 87.6 \\
Linear      & \textbf{87.1}      & \textbf{88.5} \\ \hdashline
(Learned sim.) \\
Gaussian    & 86.5      & 87.6 \\
Linear      & \textbf{86.8}      & \textbf{88.3} \\
\midrule[0.3pt]\bottomrule[1pt]
\end{tabular}
}

\end{table}

%% file: tables/toy_dataset.tex
\begin{table}[ht!]
\centering
\vspace{-3mm}

\caption{Toy Dataset: Domain attributes. Domain A, B and C are source, Domain D is target. \label{tab:toy_attributes}}

\resizebox{0.6\linewidth}{!}{
\begin{tabular}{l*{2}{c}}
\toprule[1pt]\midrule[0.3pt]
\textbf{Channel with signal}    & \multicolumn{2}{c}{\textbf{Noise std. dev.}} \\ \cmidrule(lr){2-3} 
                                & 2 & 4 \\ \midrule
Channel 1                       & A & B \\
Channel 1 and Channel 2         & C & D \\
\midrule[0.3pt]\bottomrule[1pt]
\end{tabular}
}

\vspace{-3mm}
\end{table}

%% file: tables/target_risk_full.tex
\begin{table}[ht!]
\centering

\caption{Toy Dataset: Comparison of bound terms and target risk. \label{tab:target_risk_full}}

\resizebox{\linewidth}{!}{
\begin{tabular}{*{9}{l}c}
\toprule[1pt]\midrule[0.3pt]
\textbf{Method} & \multicolumn{4}{c}{\textbf{Source Risk}}  & \multicolumn{4}{c}{\textbf{$\mathcal{H}$-divergence}} & \textbf{Target Risk} \\ \cmidrule(lr){2-5} \cmidrule(lr){6-9} \cmidrule(lr){10-10}
                & A     & B     & C     & Avg   & A     & B     & C     & Avg       & D \\ \midrule     
DANN-DG            & 0.207 & 0.375 & 0.005 & 0.196 & 1.992 & 0.800 & 2     & 1.597     & 0.399 \\
CDANN-DG           & 0.242 & 0.352 & 0     & 0.198 & 1.972 & 0.800 & 2     & 1.591     & 0.333 \\
Ours            & 0.157 & 0.362 & 0.070 & 0.196 & 2     & 0.608 & 2     & 1.536     & 0.111 \\
\midrule[0.3pt]\bottomrule[1pt]
\end{tabular}}

\end{table}

%% file: tables/hyperparameter.tex
\begin{table*}[htb]
\centering
\caption{Setup for hyperparameter tuning. \label{tab:hyperparameter}}

\begin{tabular}{*{3}{l}}
\toprule[1pt]\midrule[0.3pt]
\textbf{Method} & \textbf{Hyperparameter}   & \textbf{Distribution}   \\ 
\midrule
IRM             
& \multirow{2}{*}{\parbox{5cm}{Regularization $\lambda$ \\ Iterations of penalty annealing}}
& \multirow{2}{*}{\parbox{2cm}{$10^{Unif(-1,5)}$ \\ $\lfloor 10^{Unif(0,4)} \rfloor$}} \\
&\\
\midrule
GroupDRO        & Group weight temperature $\eta$   & $10^{Unif(-3,-1)}$ \\ 
\midrule
Interdomain Mixup 
& Beta shape parameter $\alpha$     & $10^{Unif(-1,1)}$ \\
\midrule
MTL             & Embedding averaging proportion    & \{0.5,0.9,0.99,1\}\\
\midrule
MLDG            & Meta-learning loss $\beta$        & $10^{Unif(-1,1)}$\\
\midrule
Correlation     & Regularization $\lambda$          & $10^{-5}$ \\
\midrule
CORAL-DG, MMD-DG      
& Regularization $\lambda$          & $10^{Unif(-3,-1)}$\\
\midrule
DANN-DG, CDANN-DG
& \multirow{2}{*}{\parbox{5cm}{Discriminator learning rate \\ Discriminator weight decay \\ Discriminator Adam $\beta_1$ \\ Discriminator steps \\ Discriminator gradient penalty \\ Adversarial regularization $\lambda$}}
& \multirow{2}{*}{\parbox{2cm}{$10^{Unif(-5,-3.5)}$ \\ $10^{Unif(-6,-2)}$ \\ \{0, 0.5\} \\ $\lfloor 2^{Unif(0,3)} \rfloor$ \\ $10^{Unif(-2,1)}$ \\ $10^{Unif(-2,2)}$}} \\
&\\
&\\
&\\
&\\
&\\
\midrule
VREx
& \multirow{2}{*}{\parbox{5cm}{Regularization $\lambda$ \\ Iterations of penalty annealing}}
& \multirow{2}{*}{\parbox{2cm}{$10^{Unif(-1,5)}$ \\ $\lfloor 10^{Unif(0,4)} \rfloor$}} \\
&\\
\midrule
RSC
& \multirow{2}{*}{\parbox{4cm}{Feature drop percentage $p$ \\ Batch percentage}}
& \multirow{2}{*}{\parbox{2cm}{$Unif(0,0.5)$ \\ $Unif(0,0.5)$}} \\
&\\
\midrule
Ours
& \multirow{2}{*}{\parbox{4cm}{Regularization $\lambda$ \\ RBF kernel parameter $\xi$}}
& \multirow{2}{*}{\parbox{2cm}{$10^{Unif(-3,-1)}$ \\ $ 10^{Unif(-2,2)}$}} \\
&\\
\midrule[0.3pt]\bottomrule[1pt]
\end{tabular}

\end{table*}

%% file: tables/networks.tex
\begin{table*}[htb]
\centering
\caption{Backbone network architectures for each dataset. Convolution operation is abbreviated as `Conv' and fully connected operation is abbreviated as `FC'. \label{tab:networks}}

\subfloat[Network for Bearings]{
\begin{tabular}{*{3}{l}}
\toprule[1pt]\midrule[0.3pt]
\textbf{Layer} & \textbf{Operation}   & \textbf{Specifications}   \\ 
\midrule
Convolution             
& \multirow{3}{*}{\parbox{1.5cm}{Conv \\ BatchNorm \\ LeakyReLU}}
& \multirow{3}{*}{\parbox{5cm}{8 (filter: $64\times 1$, stride: 2, pad: 1) \\ \\}} \\
&\\
&\\
\midrule
\multirow{3}{*}{\parbox{2.5cm}{Convolution \\ (3 times) \\}}
& \multirow{3}{*}{\parbox{1.5cm}{Conv \\ BatchNorm \\ LeakyReLU}}
& \multirow{3}{*}{\parbox{5cm}{8 (filter: $3\times 8$, stride: 2, pad: 1) \\ \\}} \\
&\\
&\\
\midrule
Convolution
& \multirow{2}{*}{\parbox{1.5cm}{Conv \\ LeakyReLU}}
& \multirow{2}{*}{\parbox{5cm}{8 (filter: $3\times 8$, stride: 2, pad: 1) \\}} \\
&\\
\midrule
Convolution
& Conv
& 8 (filter: $8\times 8$, stride: 1, pad: 1) \\
\midrule
Fully connected     & FC    & 32 \\
\midrule
\multirow{2}{*}{\parbox{2.5cm}{Fully connected \\ (2 times) \\}}
& \multirow{2}{*}{\parbox{1.5cm}{FC \\ ReLU}}
& \multirow{2}{*}{\parbox{5cm}{32 \\}} \\
&\\
\midrule
Fully connected     & FC    & 10 \\
\midrule[0.3pt]\bottomrule[1pt]
\end{tabular}
}

\subfloat[Network for HHAR (hidden\_dim = 128) and MIMIC-III (hidden\_dim = 32)]{
\begin{tabular}{*{3}{l}}
\toprule[1pt]\midrule[0.3pt]
\textbf{Layer} & \textbf{Operation}   & \textbf{Specifications}   \\ 
\midrule
Convolution             
& \multirow{3}{*}{\parbox{1.5cm}{Conv \\ BatchNorm \\ LeakyReLU}}
& \multirow{3}{*}{\parbox{5.5cm}{hidden\_dim (filter: $8\times 3$, stride: 1, pad: 1) \\ \\}} \\
&\\
&\\
\midrule
Convolution             
& \multirow{3}{*}{\parbox{1.5cm}{Conv \\ BatchNorm \\ LeakyReLU}}
& \multirow{3}{*}{\parbox{5.5cm}{2*hidden\_dim (filter: $5\times \text{hidden\_dim}$, stride: 1, pad: 1) \\ \\}} \\
&\\
&\\
\midrule
Convolution             
& \multirow{3}{*}{\parbox{1.5cm}{Conv \\ BatchNorm \\ LeakyReLU}}
& \multirow{3}{*}{\parbox{5.5cm}{hidden\_dim (filter: $3\times \text{(2*hidden\_dim)}$, stride: 1, pad: 1) \\ \\}} \\
&\\
&\\
\midrule
Pooling & Average pooling   & 1 (filter: 121, stride:121)\\
\midrule
Fully connected     & FC    & 6 \\
\midrule[0.3pt]\bottomrule[1pt]
\end{tabular}
}

\end{table*}

%% file: tables/datasets_samplesize.tex
\begin{table*}[htb]
\centering
\caption{HHAR: Sample size distribution per domain. \label{tab:datasets_samplesize}}

\adjustbox{max width=0.7\linewidth}{
\begin{tabular}{lp{0.95cm}p{0.95cm}p{0.95cm}p{0.95cm}p{0.95cm}p{0.95cm}}
\toprule[1pt]\midrule[0.3pt]
\textbf{Domain}   & \multicolumn{6}{c}{\textbf{Class}}  \\ \cmidrule{2-7}
    & Biking    & Standing  & Sitting   & Walking   & Stair down    & Stair up\\ \midrule
A   & 626       & 933       & 652       & 676       & 874           & 778\\
B   & 346       & 468       & 316       & 341       & 435           & 376\\
C   & 175       & 234       & 162       & 176       & 207           & 212\\
D   & 298       & 237       & 264       & 226       & 223           & 275\\
E   & 999       & 682       & 681       & 771       & 692           & 1013\\
F   & 487       & 387       & 312       & 407       & 370           & 495\\
G   & 234       & 196       & 161       & 199       & 186           & 245\\
H   & 385       & 251       & 267       & 298       & 253           & 331\\
I   & 539       & 817       & 628       & 768       & 723           & 857\\
J   & 293       & 427       & 312       & 374       & 358           & 445\\
K   & 147       & 213       & 168       & 211       & 164           & 244\\
L   & 275       & 229       & 264       & 265       & 248           & 300\\
\midrule[0.3pt]\bottomrule[1pt]
\end{tabular}}

\end{table*}